# Design and Characterization of a Micro-Vibration Adhesion System

Siqian Li
*Rolls-Royce UTC in Manufacturing and On-Wing Technology,Faculty of Engineering*
*University of Nottingham*
Nottingham,UK
Siqian.Li@nottingham.ac.uk

Xi Wang*
*Rolls-Royce UTC in Manufacturing and On-Wing Technology,Faculty of Engineering*
*University of Nottingham*
Nottingham,UK
xi.wang1@nottingham.ac.uk
Corresponding author

Jung-Che Chang
*Rolls-Royce UTC in Manufacturing and On-Wing Technology,Faculty of Engineering*
*University of Nottingham*
Nottingham,UK
j.chang@nottingham.ac.uk

Xin Dong
*Rolls-Royce UTC in Manufacturing and On-Wing Technology,Faculty of Engineering*
*University of Nottingham*
Nottingham,UK
xin.dong@nottingham.ac.uk

***Abstract*—In recent years, miniature wall-climbing robots have attracted widespread attention due to their significant potential in equipment inspection and in-situ repair applications. Traditional wall-climbing systems typically rely on electromagnetic, electrostatic, vacuum suction, or van der Waals forces for controllable adhesion. However, these conventional methods impose limitations when striving for both a compact design and high-speed mobility. This paper proposes a novel Vibration-Based Adhesion (VBA) technique, which utilizes a flexible disk vibrating near a surface to generate a strong and controllable attractive force without direct contact. By employing an electric motor as the vibration source, the constructed VBA system was experimentally evaluated, achieving an adhesion-to-weight ratio exceeding 51 times. The experimental results demonstrate that this adhesion mechanism not only provides a high normal force but also maintains minimal shear force, making it particularly suitable for high-speed movement and heavy load applications in miniature wall-climbing robots.**

## I. INTRODUCTION

The adaptability and traversal capabilities of robots significantly influence their applications in industrial inspection, infrastructure maintenance, and environmental monitoring[1]. In order to operate within complex spatial environments, continuum robots [2][3] and miniature crawling robots [4][5][6] have been developed. However, these robots are restricted by limited travel distances and the ubiquitous presence of vertical or inverted surfaces, making miniature wall-climbing robots the ultimate solution for work in structured spaces.

A critical challenge in the design of these systems is achieving a stable and controllable adhesion mechanism while maintaining a compact size and high mobility. Controllable adhesion capacity, the ability to selectively attach or detach from a surface, is a fundamental requirement within various designed systems [7]. This technique is complemented by engineered systems, such as wall-climbing robots [8] [9]and gripper [10][11]. It can be beneficial in multiple areas, including the upkeep and examination of extensive infrastructure, tracking, observing, and analyzing conditions unsuitable for human presence [9][12] [13].

Numerous adhesion methods have been developed, including active pneumatic suction [7] [14] [15], electromagnetic adhesion [16][17], bio-inspired dry/wet adhesion [18] [19], thermal adhesion [20][21] and electrostatic adhesion [9][22]. These methods each have distinct advantages, disadvantages, and specific application scopes:

- **Active Pneumatic Suction:** This method creates a negative pressure on the adhesion surface but usually requires bulky pneumatic components and is energy-intensive.
- **Electromagnetic Adhesion:** Effective only on ferromagnetic surfaces, limiting its application.
- **Bio-Inspired Dry/Wet Adhesion:** Inspired by biological adhesion structures, this method works on a variety of surfaces but is prone to performance degradation due to surface contamination.
- **Thermal (Chemical) Adhesion:** Relies on thermal or chemical reactions to generate adhesion; however, its high energy demand and slow response are major drawbacks.
- **Electrostatic Adhesion:** Features a simple structure, high energy efficiency, and minimal damage to the tested surface. Nevertheless, the slow release of the accumulated charge limits its operating speed.

Traditional adhesion methods each offer advantages, but they often suffer from high energy consumption, limited applicable surfaces, and slow response times(Table Ⅰ) [1]. Recently, the emergence of vibration-based adhesion (VBA) provides a promising alternative.

The VBA technique employs a flexible vibrating disk to form a thin air film between the disk and the target surface, thereby generating adhesion through air dynamic effects. This method not only avoids direct contact, which may cause surface contamination or damage, but also enables rapid adhesion and detachment through the adjustment of vibration frequency and centrifugal force. Studies have shown that when the vibration frequency reaches a certain threshold, the adhesion force increases linearly, providing a solid theoretical basis for designing high-performance vibration adhesion systems. Thus, VBA shows significant promise in improving adhesion efficiency, reducing energy consumption, and enhancing response speed in applications such as non-destructive testing and in-situ repair.

TABLE I. COMPARISON OF DIFFERENT ADHESION METHODS

| Requirements | Magnetic | Suction cups | Chambers | Vortex | Claws | Grips | Electrostatic | Chemical |
|---|---|---|---|---|---|---|---|---|
| *Materials* | - | o | + | + | + | - | + | o |
| *Roughness* | + | - | o | + | + | + | + | o |
| *Payload* | + | - | + | o | - | o | o | - |
| *Reliability* | + | - | o | o | + | + | + | - |
| *Forces* | + | o | + | o | + | + | + | - |
| *Consumables* | + | o | o | + | o | o | + | - |

## II. METHODOLOGY

### A. Overall System Design

The VBA system designed in this study mainly comprises the following components:

- **Vibration Source:** An electric motor (ERM motor) is used as the vibration actuator, which generates periodic vibrations by rotating an eccentric mass.
- **Flexible Vibrating Disk:** Made of PET, the disk is designed with a radius of 6.93 cm and a thickness of 2.54 mm to achieve optimal adhesion performance.
- **Connecting Adapters:** Acrylic adapters, precisely fabricated via laser cutting and connected by struts, secure the motor to the flexible disk, ensuring the stability of the overall structure.
- **Force Measurement Setup:** A load cell is used to directly measure the adhesion and centrifugal forces generated by the vibration, thus quantitatively evaluating the system's performance.

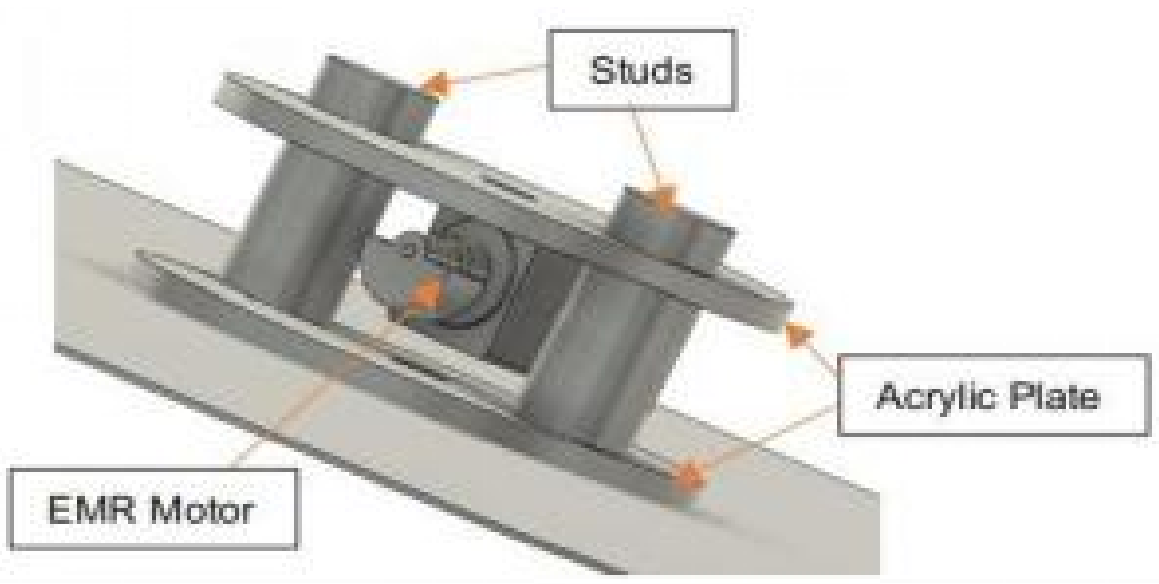

Fig. 1. CAD Model of Vibration Source

### B. Vibration Source Design and Implementation

As the core of the system, the design and regulation of the vibration source are critical to the performance of the VBA mechanism. In the experimental setup, the ERM motor, equipped with an eccentric mass, is used to generate centrifugal vibrations. The CAD model (see Fig. 1) illustrates the design, which ensures proper interaction between the motor and the flexible disk. During operation, the motor-induced vibration causes the disk to oscillate continuously, forming a thin air layer between the disk and the test surface that facilitates adhesion. For initial testing, a smaller disk with a radius of 1.75 cm and a thickness of 1.01 mm was used for convenience.

### C. System Assembly and Preliminary Testing

During assembly, the acrylic adapters were first fabricated via laser cutting. The ERM motor was then attached to the upper adapter using hot-melt glue and an off-center nut, while the flexible disk was mounted onto the lower adapter. Subsequent tests involved adjusting the motor's vibration frequency. The experimental results showed that when the motor was active, the system successfully generated sufficient adhesion to lift a lightweight acrylic board; when the motor was turned off, the adhesion force rapidly diminished, causing the board to detach and fall (see Fig. 2). These observations validate the operational feasibility of the VBA mechanism in meeting the basic load requirements of wall-climbing robots.

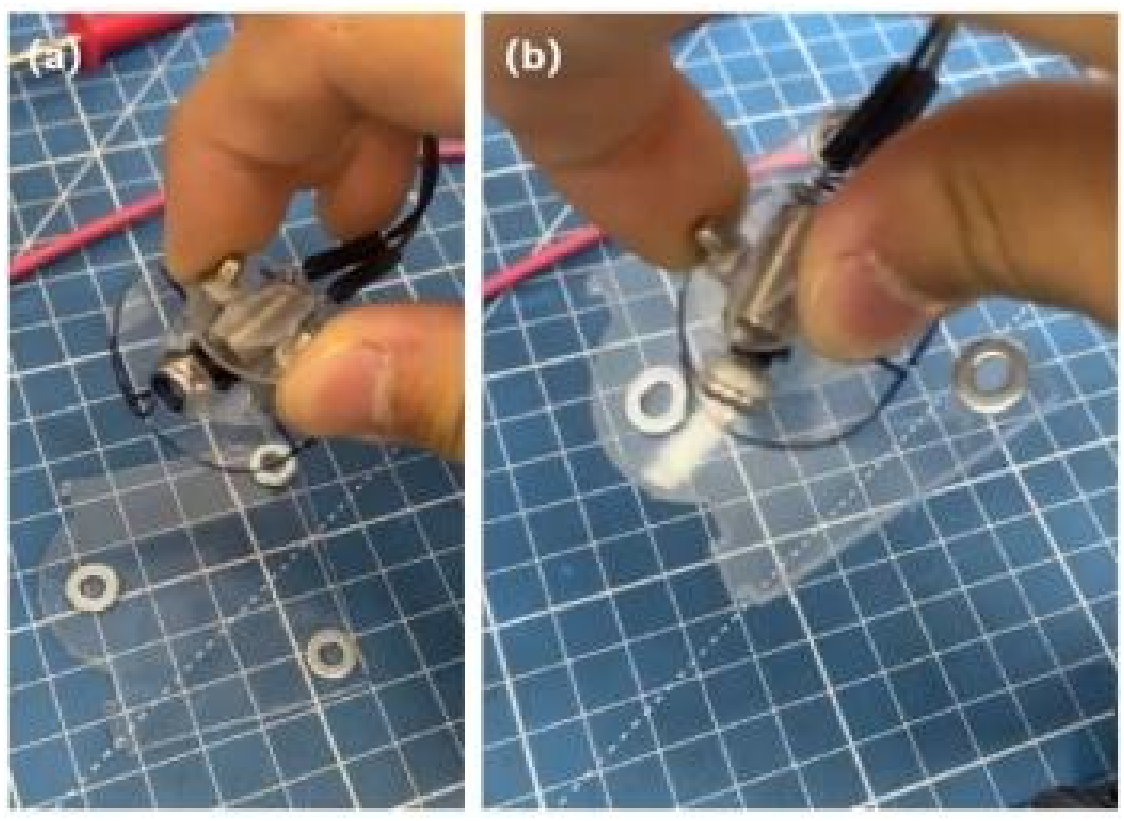

Fig. 2. (a) With the EMR Motor Off (unable to attract and lift the acrylic board); (b) With the EMR Motor On (able to attract and lift the acrylic board)

### D. Measurement of Adhesion and Centrifugal Forces

Due to the open-loop control of the ERM motor, an encoder could not be used to directly measure the vibration frequency. An alternative method, similar to sound frequency measurement, was adopted to determine the relationship between vibration frequency and adhesion force. The test setup showed in Fig. 3 involved fastening a slender rope between the load cell and the VBA device, with a fixed pulley converting horizontal tension to vertical tension for direct reading of the adhesion force.

For measuring the centrifugal force, the motor was directly attached to the load cell. Data were collected over a frequency range from 100 Hz to 200 Hz, allowing the establishment of a correlation between the frequency and the centrifugal force output.

## III. Experimental Results and Data Analysis

### *A. Effect of Vibration Frequency on Adhesion Force*

Experimental data indicate that at lower vibration frequencies (<140 Hz), the system produces insufficient adhesion force for practical applications (Fig. 4). As the frequency increases, the air film effect induced by the vibrating disk is enhanced, leading to a significant increase in adhesion performance. Specifically, once the frequency exceeds 140 Hz, a clear linear trend in the adhesion force is observed. This finding supports the notion that increasing the vibration frequency can effectively improve the overall adhesion performance, thus guiding the optimization of the vibration source design.

### *B. Results of Centrifugal Force Measurements*

Experimental data indicate that at lower Centrifugal force, serving as a critical indicator of the output of the vibration source, was also found to be directly proportional to the vibration frequency. Testing within the 100–200 Hz range revealed that the centrifugal force increases rapidly at higher frequencies. Notably, when the frequency surpasses 140 Hz, the load cell records a centrifugal force exceeding 1N, confirming that the system can provide adequate adhesion for supporting heavier loads in high-frequency operations. This characteristic is particularly advantageous for wall-climbing robots that require both rapid movement and significant load-bearing capacity.

### *C. Conclusions of Results*

The experimental results confirm that the VBA system can achieve an adhesion-to-weight ratio exceeding 51 times, providing strong theoretical and practical support for the application of this technology in miniature wall-climbing robots. It is important to note that due to the open-loop control scheme, some measurement errors in vibration frequency may exist. Future research may benefit from incorporating closed-loop control systems to enhance precision. Furthermore, the relationship between centrifugal force and adhesion force under various loading conditions warrants further investigation to comprehensively assess the stability and reliability of the adhesion mechanism in real-world applications.

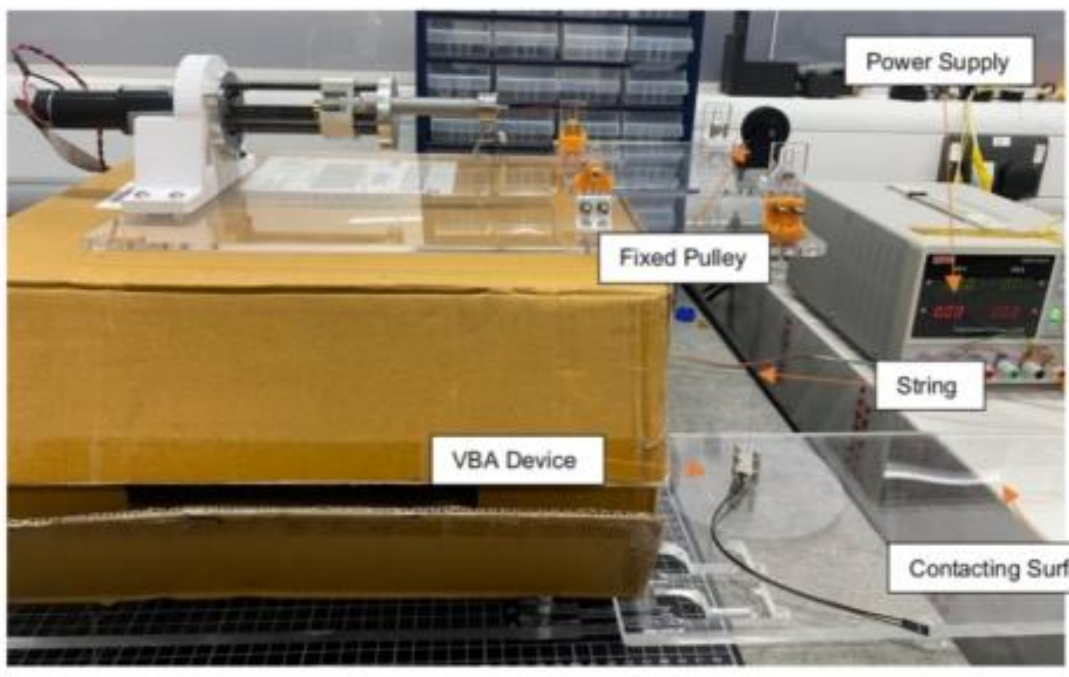

Fig. 3. Experimental Setup for Testing VBA Vibrating Frequency

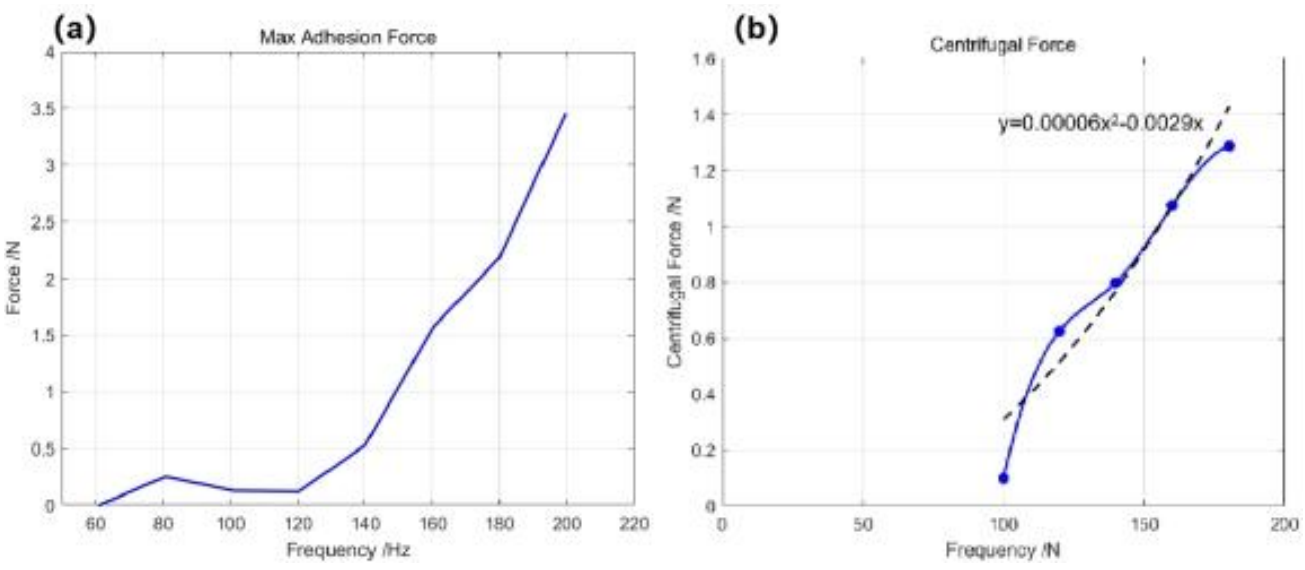

Fig. 4. (a) Frequency vs. Adhesion Force; (b) Frequency vs. Centrifugal Force

## IV. Discussion

Experimental data indicate that at lower vibration frequencies (<140 Hz), the system produces insufficient adhesion force for practical applications. As the frequency increases, the air film effect induced by the vibrating disk is enhanced, leading to a significant increase in adhesion performance. Specifically, once the frequency exceeds 140 Hz, a clear linear trend in the adhesion force is observed. This finding supports the notion that increasing the vibration frequency can effectively improve the overall

This study presents the design and construction of a vibration-based adhesion system for miniature wall-climbing robots. Experimental results demonstrate that by using an ERM motor as the vibration source and by regulating both the vibration frequency and centrifugal force, it is possible to achieve effective adhesion without direct contact with the testing surface. The primary conclusions of this study are as follows:

- **Feasibility of the System Structure:** The integration of a flexible disk with motor-induced vibrations effectively generates immediate adhesion upon activation, thereby meeting the basic load requirements for lifting lightweight objects.
- **Relationship Between Adhesion Force and Frequency:** Data indicate that adhesion force exhibits a clear linear increase when the vibration frequency exceeds 140 Hz, with a corresponding rise in centrifugal force, ensuring the system's capability for high-load applications.
- **Technical Advantages:** Compared to traditional adhesion methods, the VBA technique offers significant benefits in terms of compactness, rapid response, and energy efficiency, making it a promising approach for non-destructive testing and in-situ repair tasks.

Nevertheless, while the feasibility of the VBA technology has been demonstrated in a laboratory setting, several challenges remain for practical engineering applications:

- **Enhanced Vibration Control Accuracy:** Future studies should explore the implementation of closed-loop control and high-precision sensors to achieve more accurate regulation of vibration frequency and amplitude.
- **Multi-Environment Adaptability Testing:** Further research is required to investigate the adhesion performance on surfaces with varying material properties and roughness to ensure consistent performance under diverse conditions.

- **Optimization of Energy Consumption and Durability:** Reducing energy consumption and extending component lifespan during prolonged operation remain key issues to address.

In conclusion, this study offers a novel technical approach for the adhesion mechanism of miniature wall-climbing robots based on vibration-induced adhesion, revealing promising prospects for rapid and efficient adhesion in various practical applications. With further system optimization and rigorous testing, this technology may significantly advance the fields of intelligent inspection and repair.